# What deep learning can tell us about higher cognitive functions like mindreading?

Jaan Aru & Raul Vicente

Humboldt-University of Berlin, Germany
Institute of Computer Science, University of Tartu, Estonia

Jaan Aru, jaan.aru@gmail.com
Raul Vicente, raulvicente@gmail.com

**Abstract**

Can deep learning (DL) guide our understanding of computations happening in biological brain? We will first briefly consider how DL has contributed to the research on visual object recognition. In the main part we will assess whether DL could also help us to clarify the computations underlying higher cognitive functions such as Theory of Mind. In addition, we will compare the objectives and learning signals of brains and machines, leading us to conclude that simply scaling up the current DL algorithms will most likely not lead to human level Theory of Mind.

# Introduction

In this perspective we will first very briefly review how deep learning (DL) has helped us to study visual object recognition in the primate brain (see Kriegeskorte, 2015; Yamins & DiCarlo, 2016 for more thorough reviews). The successful application of DL in vision leads to the question whether these models could help us to also gain insights into other aspects of human cognition. It is presently unclear whether DL can lead to improved understanding of other cognitive processes beyond vision. In the main part of the paper we ask whether DL could also help us to understand the emergence of higher cognitive functions such as Theory of Mind.

# Marrying biology and AI: The success story in vision

## DL as a model for primate vision

Visual object recognition in primates is mediated by a hierarchy of transformations along the occipitotemporal cortex (DiCarlo et al., 2012). Intriguingly, it has been shown that these transformations are quite similar to the hierarchy of transformations learned by deep neural networks (DNN) trained to recognize objects on natural images. Several pieces of work have demonstrated a direct correspondence between the hierarchy of the human visual areas and the layers of the DNNs (Gülcu 2015, Seibert, 2016, Cichy et al., 2016; Eickenberg 2016; Kuzovkin et al., 2018).

While DL has offered an algorithmic model for (feedforward) visual object recognition in brains, these developments do not provide a full understanding of biological vision (Cox, 2014; Kriegeskorte, 2015; VanRullen, 2017). Most importantly, in biological vision at least part of the processing is done by feedback connections (Roelfsema, 2006), although the exact computational role of feedback is less clear (See Bastos et al, 2012 for one particular view). The DL networks commonly used in machine vision are feedforward, although there is a recent trend towards incorporating feedback (e.g. Wen et al., 2018). Also, it is important to note that the DL networks still explain only a part of the variability of the neural responses happening in real brains. Hence, the present-day DL networks cannot be seen as the ultimate model of biological visual processing (e.g. Kriegeskorte, 2015; Rajalingham et al, 2018). Nevertheless, the work with DL has illuminated how relatively simple transformations applied throughout a hierarchy of processing stages can be associated with successful object recognition.

## What has made DL successful for investigating biological vision?

The recent developments in DL have been made possible by increased computational power, refinements to the algorithms, and availability of large data-sets necessary for training DL networks (LeCun et al., 2015). Beyond these factors there are several specific aspects to consider that have enabled DL to be helpful for investigating biological vision. There are three aspects to highlight: 1) Appropriate training data: DNNs are a good model for biological object recognition as they are trained on datasets that are directly relevant for biological vision: natural images (i.e. Imagenet (Deng et al., 2009)); 2) A training objective that is similar to biological vision: Machine vision has the straightforward goal to accurately recognize objects in a scene (or to segment or localize them) which resembles the goal of the biological vision (see below for a more nuanced view); 3) Good neuroscientific comparison data: It has become much more straightforward to measure the activity of biological vision systems. In the case of vision we have a fairly good understanding of the areas involved in vision, how they are connected and organized. Decades of work on the visual system have revealed the visual processing hierarchy which can and has been compared to the hierarchy of transformations learned by DL systems (Gülcu 2015, Seibert, 2016, Cichy et al., 2016; Eickenberg 2016). In the case of vision all these aspects together were needed for DL to inform us about the mechanisms underlying visual object recognition.

# Challenges ahead: using DL to study other aspects of cognition

Studying vision with the help of DL seems justified - both the artificial and the biological visual systems solve a similar task with a similar performance, both are hierarchical and require the transformation of features from simple to more complex. From this perspective, using DL to investigate the computations underlying mindreading might seem a bit far-fetched as Theory of Mind seems quite different from vision: Theory of Mind (or mindreading) is an essential ability of humans to infer the mental states of others such as for example their perceptual states, beliefs, knowledge, desires, or intentions (for review Apperly, 2010). While we share basic visual processes with most other mammals, it is thought that the scope and complexity of mindreading skills sets us apart from most of the animal kingdom (Call & Tomasello, 2008). Hence even if one would agree that DNNs are useful for understanding vision, it is unclear whether DNNs have anything to tell us about mindreading. How could this ability to understand what others think and intend emerge from artificial neural networks (Lake et al., 2017; Baker et al., 2017)? On the other hand, diverse research supports the view that mindreading to a large extent is an acquired skill, just like reading (Apperly, 2010; Heyes & Frith, 2014). And if mindreading skills require training, similarly to vision, then DL networks could help to unravel at least some of the computations underlying mindreading. To highlight this aspect of being learned we will henceforth use the term "mindreading" instead of "theory of mind" in this manuscript (see also Apperly, 2010; Heyes & Frith, 2014).

We feel that mindreading will be an important topic to study with the modern tools offered by DL for two several reasons. First, if the goal is to build artificial agents that think and behave at (least at) the level of humans, then there might not be a way around studying mindreading. This is because at least according to some prominent views about communication, mindreading is necessary for the emergence of meaningful communication and language (Tomasello, 2010; 2014; Scott-Phillips 2014; Mercier & Sperber, 2017). This perspective suggests that training agents with DL on huge text corpora will never lead to agents that are able to communicate with humans or with each other in a meaningful fashion. Hence, the only way to build these agents is to first understand and build in mindreading capabilities. However, this is a daunting task, as there is still much unclarity about mindreading in humans and animals (e.g. Siegal, 2008; Call & Tomasello, 2008; Apperly, 2010; Heyes & Frith, 2014; Scott & Baillargeon, 2017). Second, given that there is still much controversy about mindreading even in humans, we believe that modern DL tools can actually help to better understand mindreading. This is because in artificial systems one can add and modify single components of the system. For example, one could see how having an external memory (i.e. separately from the deep neural network) can help the agents in acquiring basic mindreading skills. Similarly, as discussed in the next section, one can see whether and which mindreading skills can emerge through reinforcement learning. Third, similarly to how we require mindreading skills to understand other humans, we will need mindreading skills in AI systems for them to understand human intentions, for example in human-machine interaction (see Rabinowitz et al., 2018). For these reasons we think there will be a surge of DL works into mindreading.

In this perspective article we want to discuss how DL could contribute to studying mindreading. We will first delineate some problems in studying mindreading with DL. Next we will offer one particular way for studying mindreading with DL. We will conclude that the present day DL algorithms are not sufficient for acquiring human like mindreading skills and describe some of these learning signals relevant for acquiring mindreading skills in biological brains.

## DL to study the emergence of mindreading

As noted above, progress in machine vision has been so beneficial for the study of biological vision because of comparable training data and similar training objectives for biological and machine vision algorithms. On top of that in the case of vision there is appropriate neuroscientific data to compare to the outcomes of DNN. All of these aspects are yet missing or severely underdeveloped in mindreading research. What would constitute good training data for teaching artificial agents how to read other minds? Which neuroscientific data could one compare the artificial models to? And, most importantly, what would be the training objective?

To make any meaningful comparison between the AI agents and biological agents with regard to mindreading, one needs to find a task that is similar or at least comparable to both (as in vision, where often the task for both biological and artificial agents is to recognize objects). Although there are several classic tasks for studying human mindreading (e.g. the Sally and Anne task), they depend on language and executive functions (see Apperly, 2010, for a review). However, we believe that as a first step it is necessary to study the core mindreading skills possessed already by preverbal infants and non-human animals (Baillargeon et al., 2010; Call & Tomasello, 2008).

In the early 2000s Brian Hare, Josep Call and Michael Tomasello performed an excellent series of experiments demonstrating that chimpanzees exhibit at least some characteristics of mindreading (Hare et al., 2000; 2001). The experiments involved two chimpanzees, a dominant (D) and a subordinate (S) chimpanzee. (Chimpanzee social status is organized hierarchically, i.e. some animals are more dominant than others.) When the same piece of food is available for both D and S, the dominant almost always obtains it. In the experiments the two chimpanzees were set into separate cages facing each other. Between them, there was a space containing two walls. During the experiment, pieces of food were presented. In the critical condition one piece of food could only be seen by the subordinate and not by the dominant chimpanzee. Could the subordinate take advantage of the fact that the dominant chimpanzee could not perceive one piece of food? The results demonstrated that the S chimpanzee indeed obtained more food in this condition. Hence, the S chimpanzee was able to take into account what the D chimpanzee could and could not see: One chimpanzee could take the perspective of another chimpanzee.

This was an elegant experiment as it demonstrated that the chimpanzees possess at least some basic form of mindreading (Call & Tomasello, 2008). Importantly, such competitive settings are easy to implement with AI agents because in this task there is a clear goal to optimize for: obtaining food. Of course we do not think that "obtaining food" is the sole goal driving mindreading skills in primates (see next section), but having such a clear goal provides a tangible starting point for studying mindreading with deep reinforcement learning (DRL), where the DNN learns through rewards (e.g. points, Mnih et al., 2015; Silver et al., 2016). We believe that implementing this task with AI agents provides a way to study the computations underlying mindreading.

In our work (Labash et al., 2019), we implemented two agents ("subordinate" and "dominant") who were competing for reward. We investigated whether the behavior of the artificial agents reveals some rudimentary skills of perspective taking, similar to the chimpanzee work (Hare et al 2000; 2001). The behavior of the agents indeed showed evidence for basic perspective taking skills, which demonstrates that at least part of perspective taking skills can indeed be learned through DRL.

Figure 1 illustrates how a subordinate agent solves the basic task: go to food if the dominant is not observing; avoid the food when the dominant is observing. As can be seen from Figure 1 the subordinate agent still occasionally performs the unexpected behavior, but all in all it has learned to behave as if it would take into account what the dominant can or can not see.

It is important to note that we are not claiming that DRL would capture all aspects of mindreading. However, by understanding the capabilities and limitations of DRL in acquiring mindreading we will better understand the computational demands of mindreading, just as DNNs have led to a better understanding of vision.

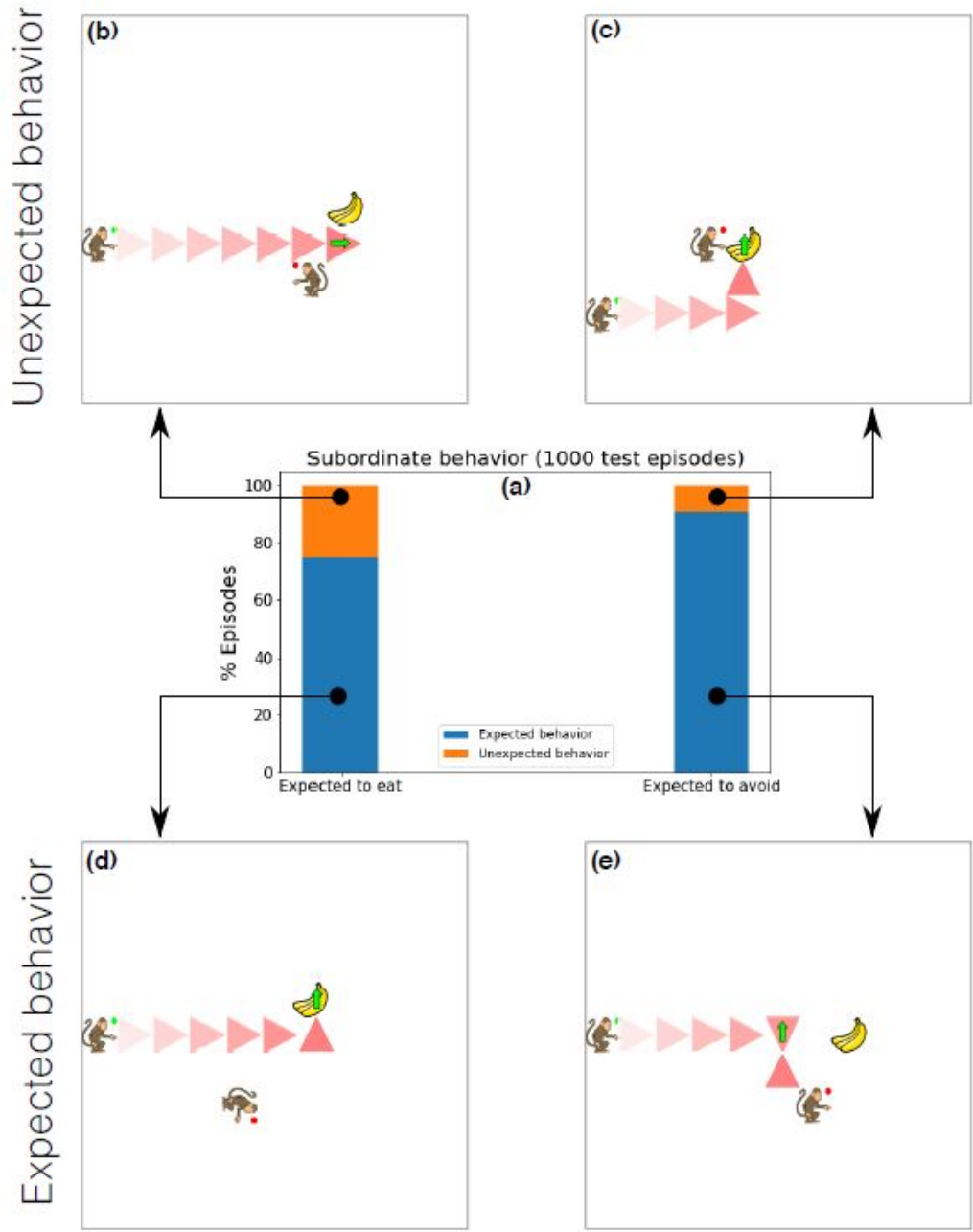

**Figure 1:** Quantification of the subordinate behavior and examples of model trajectories **a)** Bar plot with the percentage when the model performed the expected optimal behavior. Two expected behaviors are distinguished: 1) agent is expected to eat when the food is not observed by the dominant, and 2) agent is expected to avoid when the food is observed by the dominant. **b)** An example of the subordinate agent (green circle) avoiding the food although it should approach it. **c)** An example of the model reaching the food when it should not reach it. **d)** An example of the model performing the expected behavior of navigating and obtaining the food. **e)** An example of model behavior of avoiding the food when this is observed by the dominant agent (red circle).

This task based on (Hare et al 2000; 2001) is far from ideal when it comes to comparing biological and AI agents. First, such competitive paradigms are not natural for humans. This means that comparison to human behavior is lacking and that with this particular task one cannot study more complex forms of mindreading. Also, it is hard to study the brains of awake chimpanzees, hence elucidating neural mechanisms underlying this form of mindreading is hindered. This is unfortunate as it precludes any comparison between the representations in brains and representations learned by the AI agents. As noted above, our understanding of vision has benefitted from DL exactly because one can make such direct comparisons between the representations in DNNs and in biological vision (Kriegeskorte, 2015; Yamins & DiCarlo, 2016). To make this happen in the domain of mindreading, one needs to design simpler tasks that capture mindreading skills but are not dependent on language (Scott & Baillargeon, 2017, see also Rabinowitz et al., 2018). Finally, to acquire mindreading skills the AI agents probably need to be equipped with the learning signals available to human infants. In the next section we will describe some of these learning signals relevant for acquiring mindreading skills in biological brains.

## Goals, rewards and learning in brains and machines

For training AI agents it is important to determine the goal the agent should optimize for. Hence, one key question for creating AI algorithms that could be informative about biology is "what is the goal of the respective biological system" (Cox, 2014; Marblestone et al., 2016; Scholte et al., 2017). In vision, the goal "to recognize objects" could be a good proxy for what the ventral visual stream is optimized for (Yamins & DiCarlo, 2016), although it is certainly not the only goal of biological vision (Cox, 2014; Scholte et al., 2017). "Recognizing objects" is a goal that is also quite easy to implement in AI. That is one reason why AI has been very useful for the research on vision. In contrast, there is no clear goal function for training AI algorithms for mindreading skills. This is because the generic functions that mindreading skills might be optimized for (e.g. communication, deception) are themselves complex to formulate and hard to implement in AI (but for some important first steps see Foerster et al., 2016; 2017; Sukhbaatar & Fergus, 2016; Mordatch & Abbeel, 2017; Rabinowitz et al., 2018; Matiisen et al., 2018). Most likely, such higher cognitive functions arise from the combination of many different neural processes that obey their own cost functions (Marblestone et al., 2016). In the last section we offered a potential goal in the acquisition of mindreading skills: In the context of multi-agent competition (Tampuu et al., 2017), mindreading could emerge through a process of an agent trying to maximize the probability of reward (e.g. food intake) while avoiding competitive interference by other agents (Labash et al., 2019). However, we are of course not implying that maximizing the probability of food rewards is the sole goal that would lead to mindreading skills to emerge. In this section, we would like to provide a better understanding of the goals and learning signals that drive biological agents and how these differ from learning in machines.

In the present-day AI approaches it is fashionable to learn directly from data without coding prior knowledge into the network, hence "innate biases" may sound a bit like heresy (see Marcus, 2018, for a discussion about innateness in AI). However, these small biases enable the organism to learn about aspects of the world that have been important for the species over the course of evolution, not those that are the most salient, novel or statistically dominant in the current environment. By providing the agent with genetically pre-defined bias one can speed up learning the *relevant* features of its environment (Ullman et al., 2012; Marblestone et al., 2016). For example, recognizing and distinguishing other human beings is important and hence there is an innate bias for attending to faces (Johnson, 2005; Reid et al., 2017). In particular, preferences for faces over similarly configured non-face objects are present in neonatal infants (Farroni et al., 2005) and even in fetuses in the third trimester of pregnancy (Reid et al., 2017). This bias is most likely a subcortical detector for stimuli with face-like configurations (Johnson, 2005; Reid et al., 2017) that directs the attention of the organism (e.g. through an eye-movement) towards faces. These subcortical detectors bias the cortical learning system to process more input about faces and hence the organism learns faster about them (Johnson, 2005; Johnson et al., 2015).

Similar biases likely exist for drawing infant's attention to speech: they direct the learning resources of the infant to the speech signal. Speech signals are more interesting and arousing for the infant than other environmental sounds (Perszyk & Waxman, 2018) and hence lead to quicker learning about them. Further innate biases likely exist for directing the infant's attention to hands (Ullman et al., 2012), eyes and gaze-direction (Grossmann, 2017), and biological motion (Simion et al., 2008). For us humans, such innate biases are a key component of start-up software such as "intuitive psychology" (Lake et al., 2016). These biases also make humans more intrinsically motivated to learn about other humans (Oudeyer & Kaplan, 2009; Schmidhuber, 2010).

In sum, the innate part about mindreading is that over the course of human evolution external learning signals have been adjusted so that the infant brain is very sensitive to other people (Burkart et al, 2009; Hawkes, 2014): stimuli such as faces, eyes and infant-directed speech aid and guide learning in the infant brain. In other words, other people and their behavioral signals are more important for human infants than other environmental cues. We believe that this sensitivity to the behavior of others makes the organism learn more about the behavior and states of other people and hence provides one key cornerstone for developing mindreading skills. Instead of wanting to avoid any kind of innateness in AI agents, one should embrace at least some innate biases that are known to make humans smart (Marcus, 2018). Equipping AI agents with a preferential sensitivity to other agents might be necessary to even come close to human-like mindreading skills.

Our suggestion is not yet realized by the recent work of Rabinowitz and colleagues (2018). In their approach Rabinowitz et al., 2018 train a neural network to learn mindreading skills while predicting the behavior of an agent in a gridworld. Although this study represents one of the first efforts to equip artificial agents with mindreading faculties, the authors themselves acknowledge severe limitations. In their approach, the agent who was learning mindreading had full observability of the other agent and environment, it was trained in a supervised manner, and it did not act on the environment (so it can hardly be called an agent). These important aspects limit the biological plausibility of both the task and teaching signals used in the learning process. It could be argued that if training through supervision works then there is no need for "innate biases" to make the systems sensitive to other agents. However, based on our analysis we feel that whenever organisms (or artificial mindreading agents) are put to complex environments, they benefit from these biases that direct their attention and processing resources to particular features of the environment. Future research will need to show the range and limits of mindreading skills that can emerge purely through supervision.

This is not to say that supervision plays absolutely no role in animal or human learning. Social supervision shows to animals and humans "how things are done around here": when a lion cub follows her mother she learns about optimal foraging and navigation; when she observes her mother during buffalo hunt, she learns to hunt; when she performs rough and tumble play with her siblings, she learns to fight and defend. Similarly, human children observe, imitate, play and learn through these behaviors (Gopnik, 2017). Recently, the AI community has taken up the challenge to use imitation and observation of others as a teaching signal (e.g. Borsa et al., 2017; Bansal et al., 2017, Stadie et al., 2017).

Taken together, the training signals for AI agents are very primitive as compared to the teaching signals a human baby obtains from the environment. Even when the AI agent is trained by both external rewards (points) and internal rewards (prediction errors) as done in some recent works (Pathak et al., 2017, Bellemare et al., 2016), the biological learning signals are more diverse. Two conclusions follow from this discussion. First, simply scaling up the current efforts for developing AI will not lead to human level intelligence (or, more narrowly, to mindreading skills). Rather, the AI algorithms need to be supported with at least some biases in order to learn from social agents.

## Conclusions

Using DL to study visual object recognition has been a success story. Just a few years ago one could claim that nobody knows how biological visual recognition works, but now DL has provided a working model for at least some aspects of biological vision (VanRullen, 2017). DNNs offer a framework for understanding how one can go from pixels to meaningful object categorization. The success of DNNs in vision has generated a hope that DL could lead to similar progress in understanding higher cognitive functions. However, it is also possible that the case of vision is not typical: in vision there is abundant training data, good neuroscientific comparison data and a clear training objective. As a case example where it might be more complicated for DL to lead to an understanding of biological computations we considered the topic of mindreading. Mindreading "performance" does not obey a simple cost function, there is less neuroscientific data to compare to and there are no databases for training the AI agents. Nonetheless, this does not imply that AI cannot contribute to understanding the algorithms underlying mindreading in the brain. We foresee that if one includes the bias of being more sensitive to humans (and other agents) and endows the agents with other components (e.g. external memory), these agents will be more successful in mindreading tasks. Having AI agents that are closer to humans in mindreading will help us to understand how mindreading skills are learned in the human brain. Discovering which algorithmic components are necessary for acquiring mindreading abilities will contribute to resolving many debates surrounding the cognitive and biological basis of mindreading (e.g. Siegal, 2008; Call & Tomasello, 2008; Apperly, 2010; Heyes & Frith, 2014; Scott & Baillargeon, 2017).

In this perspective we have focused on how certain similarities between biological and artificial neural networks can be exploited to gain understanding on how the brain might solve some computational problems. Obviously, these systems are also different in many important respects (Lake et al., 2016). Thus, it is essential to remark that we should not take DL architectures and their learning algorithms as the ultimate brain-like learning system, but simply as working examples that can guide our search for how the brain really works. So far DL remains our best source for working algorithms in large-scale tasks similar to those that animals have to solve. Given the difficulty in monitoring all the relevant variables in real brains and the inaccessibility to the brain's masterplans, the fact that we can open these artificial algorithms and analyze them in detail provides a source of inspiration that we can not afford to not explore.

**Acknowledgements:** We thank M. Hebart, D. Majoral, T. Matiisen and A.Tampuu for helpful comments on the manuscript. This work was supported by the Estonian Research Council through the personal research grant PUT1476, and the Estonian Centre of Excellence in IT (EXCITE), funded by the European Regional Development Fund.